\documentclass{article}

\usepackage{arxiv}

\usepackage[utf8]{inputenc} 
\usepackage[T1]{fontenc}    
\usepackage{hyperref}       
\usepackage{url}            
\usepackage{booktabs}       
\usepackage{amsfonts}       
\usepackage{nicefrac}       
\usepackage{microtype}      
\usepackage{lipsum}

\usepackage{soul}
\usepackage{url}
\usepackage[utf8]{inputenc}
\usepackage{graphicx}
\usepackage{booktabs}
\usepackage{algorithm}
\usepackage{algorithmic}
\usepackage{booktabs} 
\usepackage{pifont}
\usepackage{caption}
\usepackage{float}
\usepackage{stfloats}
\usepackage{subfigure}
\usepackage{amssymb}
\usepackage{multirow}

\makeatletter 
  \newcommand\figcaption{\def\@captype{figure}\caption} 
  \newcommand\tabcaption{\def\@captype{table}\caption} 

\title{PathVQA: 30000+ Questions for Medical Visual Question Answering}

\author{
  Xuehai He \\
  University of California San Diego\\
   \And
 Yichen Zhang \\
  University of California San Diego\\
   \And
 Luntian Mou \\
  Beijing University of Technology\\
   \And
 Eric Xing \\
  Carnegie Mellon University\\
   \And
 Pengtao Xie \\
  University of California San Diego\\
}
\begin{document}
\maketitle

\begin{abstract}
Is it possible to develop an ``AI Pathologist" to pass the board-certified examination of the American Board of Pathology? To achieve this goal, the first step is to create a visual question answering (VQA) dataset where  the AI agent is presented with a pathology image together with a question  and is asked to give the correct answer. Our work makes the  first attempt to build such a dataset. Different from creating general-domain  VQA datasets where the images are widely accessible and there  are  many crowdsourcing workers available and capable of generating question-answer pairs, developing a medical VQA dataset is much more challenging. First, due to privacy concerns, pathology images are usually not publicly available. Second, only well-trained  pathologists  can understand  pathology  images, but they barely have time to help create datasets for AI research. To address these challenges, we resort to pathology textbooks and online digital libraries. We develop  a semi-automated pipeline to extract pathology images and captions from textbooks and generate question-answer pairs from captions using natural language processing. We collect 32,799 open-ended  questions from 4,998 pathology images where  each question  is manually checked to ensure correctness. To our best knowledge, this is the first dataset for pathology VQA. Our dataset will be released publicly to promote research in medical VQA. 
\end{abstract}


\section{Introduction}

Pathology studies the causes and effects of diseases or injuries. It underpins every aspect of patient care, from diagnostic testing and treatment advice to using cutting-edge genetic technologies and preventing diseases. Medical professionals practicing pathology are called pathologists, who examine bodies and body tissues. To become a board-certificated pathologist in the US, a medical professional needs to pass a certification examination organized by the American Board of Pathology (ABP), which is a very challenging task. We are interested in asking: whether an artificial intelligence (AI) system can be developed to pass the ABP examination? It is an important step towards achieving AI-aided clinical decision support and clinical education.  

\begin{figure} [t]
  \centering 
 { 
    \includegraphics[width =0.65\columnwidth]{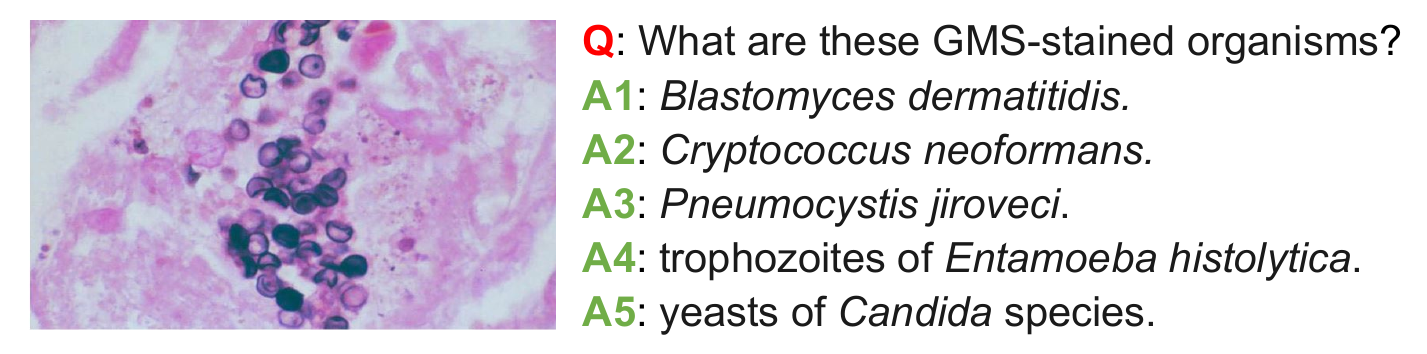}
    }
  \caption{An example of ABP test questions}
  \label{ABP}
  \end{figure}

Among the ABP test questions, one major type is to understand the pathology images. Given a pathology image and a question, the examinees are asked to select a correct answer. Figure~\ref{ABP} shows an example. To train an AI system to  pass this exam, we need to collect a dataset containing questions similar to those in the ABP test. ABP provides some sample questions, but they are too few to be useful for training data-driven models. Some commercial institutes provide a larger number of practice  questions, but they are very expensive to buy and they cannot be shared with the public due to copyright issues. 
   
To address these limitations, we aim to create a pathology visual question answering (VQA) dataset that contains questions similar to those in the ABP tests and can be shared with the broad research community on AI for healthcare. To our best knowledge, this is the first dataset for pathology VQA. VQA~\cite{vqa} is an interdisciplinary research problem that has drawn extensive attention recently. Given an image (e.g., an image showing a dog is chasing a ball) and a question  asked about the visual content of the image (e.g., ``what is the dog chasing?"), VQA aims to develop AI algorithms to infer the correct answer (e.g., ``ball"). VQA requires a deep comprehension of both images and textual questions, as well as the relationship between visual objects and textual entities, which is technically very demanding.  
%
%
%
While there have  been several datasets~\cite{first,vqa,cocoQA,clevr,goyal} for general domain VQA, datasets for medical VQA are very rare. 

It is much more challenging to build medical VQA datasets than general domain VQA datasets. First, many human workers in crowdsourcing platforms such as Amazon Mechanical Turk are available to generate questions and answers from general domain images. These images contain contents (e.g., dog, cat, lake) easily understandable to human. There is almost no barrier to comprehend the images, ask proper questions about the visual objects, and give correct answers. However, medical images such as pathology images are highly domain-specific, which can only be  interpreted by  well-educated medical professionals. It is very difficult and expensive to hire medical professionals to help create medical VQA datasets. Second, to create a VQA dataset, one first needs to collect an image dataset. While images in the general domain are pervasive, medical images are very difficult to obtain due to privacy concerns. 

\begin{figure} [t]
  \centering 
 { 
    \includegraphics[width =0.5\columnwidth]{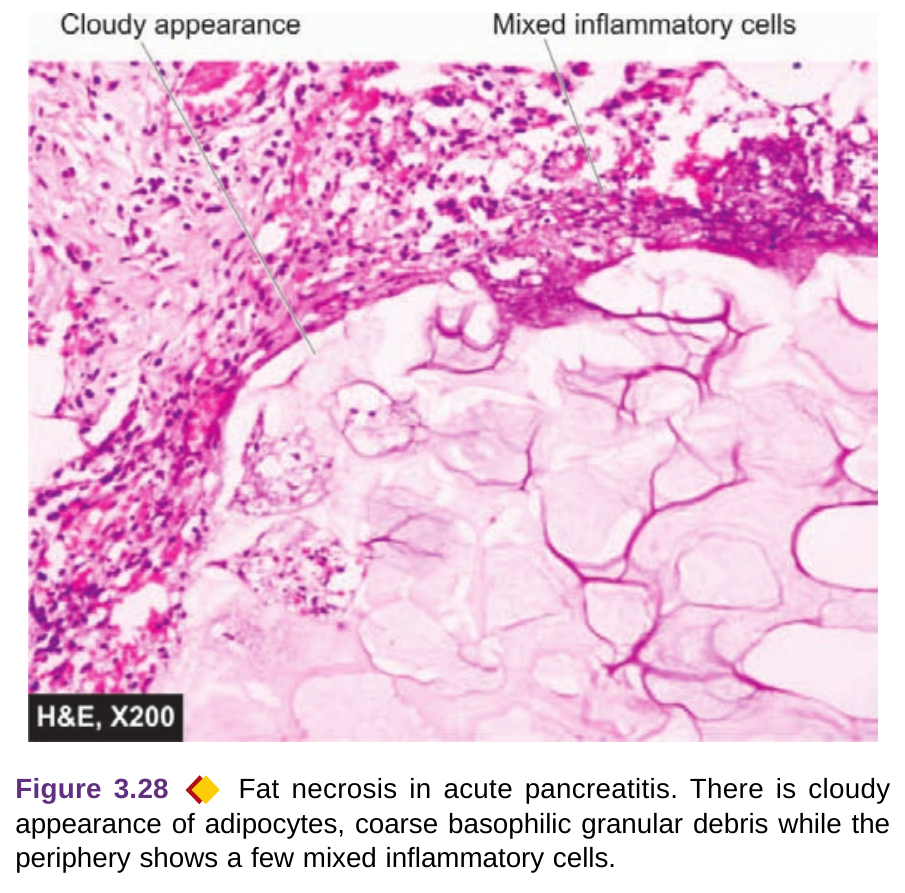}
    }
  \caption{An example of image/caption pairs from the ``Textbook of Pathology''}
  \label{caption_image_example}
  \end{figure}
   
To address these challenges, we resort to pathology textbooks, especially those that are freely accessible online, as well as online digital libraries. These textbooks contain a lot of pathology images, covering the entire domain of pathology. Each image has a caption  that describes pathological findings present in the image (as shown in Figure \ref{caption_image_example}). The caption is carefully worded and clinically precise. We extract images and captions from the textbooks and online digital libraries and develop a semi-automated pipeline to generate question-answer pairs from each caption. We have manually checked the automatically-generated questions and answers and fixed small grammatical issues. In the end, we collected a pathology VQA dataset containing 4,998 images and 32,799 question-answer pairs. 

The major contributions of this paper are as follows:
\begin{itemize}
    \item We create a pathology visual question answering (VQA) dataset containing 4998 pathology images and 32,799 question-answer pairs to foster the research of medical VQA. To our best knowledge, this is the first dataset for pathology VQA.
    \item We develop an semi-automated pipeline to efficiently create medical VQA datasets from medical textbooks and online digital libraries. Our pipeline can be widely applied to other medical imaging domains beyond pathology, such as radiology, ultrasound, etc. 
    \item We apply several well-established and state-of-the-art VQA methods to our dataset and generate a set of baseline results for  other researchers to benchmark with. 
\end{itemize}

The rest of the paper is organized as follows. Section 2 presents an overview of existing VQA datasets. Section 3 describes our pipeline for constructing pathology VQA datasets from pathology textbooks and online digital libraries. Section 4 presents the statistics of our dataset. Section 5 introduces baselines VQA models and results achieved on our dataset.  Section 6 concludes the paper.

\section{Related Works}
\subsection{Datasets}

\begin{table}[t]
    \centering
    \caption{Comparison of VQA datasets}
    \scalebox{0.9}{
\begin{tabular}{|c|c|c|c|c|}
\hline  
 &Domain&\# images& \# QA pairs &Answer type\\
\hline  
DAQUAR&General& 1,449&12,468  & Open \\
\hline
VQA&General& 204K& 614K & Open/MC \\
\hline
VQA v2 &General& 204K& 1.1M & Open/MC \\ 
\hline
COCO-QA & General&123K& 118K & Open/MC \\
\hline
CLEVR&General& 100K& 999K & Open \\ 
\hline
\hline
VQA-Med &Medical& 4,200& 15,292 & Open/MC \\ 
\hline
VQA-RAD &Medical& 315& 3,515 & Open/MC \\ 
\hline
Ours&Medical& 4,998& 32,799 & Open \\ 
\hline
\end{tabular}}

\label{dataset}
\end{table}

To our best knowledge, there are two existing datasets for medical visual question
answering.
 The VQA-Med~\cite{medicalVQA_dataset19} dataset is created on 4,200 radiology images and has 15,292 question-answer pairs. There are four categories of  clinical questions: modality, plane, organ  system, and abnormality. For the first  three categories, the QA is  in multiple-choice (MC) style where the number of possible  answers is fixed (36, 16, and 10 respectively). Consequently, the QA tasks can be equivalently formulated as multi-way classification problems with 36, 16, and 10 classes respectively. This makes the difficulty of this dataset significantly lower. Questions in the abnormality category are truly challenging open-ended questions. However, there are only 2408 such questions (15.7\%). VQA-RAD~\cite{nature} is a manually-crafted dataset where questions and answers are given by clinicians on radiology images. It has 3515 questions of 11 types, e.g. modality, plane, etc. 58\% of the questions are in MC style and the rest are open-ended.
Our dataset  differs from VQA-Med and VQA-RAD in two-fold. First, our dataset is about pathology while VQA-Med and VQA-RAD~\cite{nature} are both about radiology. Second, our dataset is a truly challenging QA dataset where most of the questions are open-ended while in VQA-Med and VQA-RAD the majority of questions have a fixed  number of  candidate answers and can be answered by multi-way classification. Besides, the number of questions in our dataset is much larger than that in VQA-Med and VQA-RAD.

A number of visual question answering datasets have been developed in the general domain. DAQUAR~\cite{first} is built on top of the NYU-Depth V2 dataset~\cite{nyu} which contains RGBD images of indoor scenes. DAQUAR consists of (1) synthetic question-answer pairs that are automatically generated based on textual templates and (2) human-created question-answer pairs produced by five annotators. The VQA dataset~\cite{vqa} is developed on real images in MS COCO~\cite{coco} and abstract scene images
in~\cite{abstract0,abstract1}. The
question-answer pairs are created by human annotators who are encouraged to ask ``interesting" and ``diverse" questions. VQA v2~\cite{goyal} is extended from the VQA~\cite{vqa} dataset to achieve more balance between visual and textual information, by collecting complementary images in a way that each question is associated with a pair of similar images with different answers. In the COCO-QA~\cite{cocoQA} dataset, the  question-answer pairs are automatically generated from image captions based on syntactic parsing and linguistic rules. CLEVR~\cite{clevr,man} is a dataset developed on rendered images of spatially related objects (including cube, sphere, and cylinder)  with different sizes, materials, and colors. The locations and attributes of objects are annotated for each image. The questions are  automatically generated from the annotations. 

Table~\ref{dataset} presents a comparison of different VQA datasets. The first five datasets are in the general domain while the last three are in the medical domain. Not surprisingly, the size of general-domain datasets (including the number of images and question-answer pairs) is much larger than that of medical datasets since general-domain images are much more available publicly and  there are many qualified human annotators to generate QA pairs on general images.

\subsection{Automatic Construction of Question-Answer Pairs}
Existing datasets have used automated methods for constructing question-answer pairs. In DAQUAR, questions are generated with templates, such as ``How many \{object\} are in \{image\_id\}?". These templates are instantiated with ground-truth facts from the database. In COCO-QA, the authors develop a question generation algorithm based on the Stanford syntactic parser~\cite{stanfordParser}, and they form four types of questions\textemdash``object", ``number", ``color", and ``location" using hand-crafted rules.    
In CLEVR, the locations and attributes  of objects in  each image are fully annotated, based on which the questions  are generated by an automated algorithm. Their algorithm cannot be  applied to natural  images where detailed annotation of objects and  scenes are very difficult to obtain. In~\cite{auto_question_generation}, the authors develop a conditional auto-encoder~\cite{VAE} model to automatically generate questions from images. To train such a model, image-question pairs are needed, which incurs a chicken-and-egg problem: the goal is to generate questions, but realizing this goal needs generated questions.  
%
In VQA-Med, the authors collect medical images along with associated side information (e.g., captions, modalities, planes) from the MedPix\footnote{https://medpix.nlm.nih.gov} database and generate question-answer pairs based on manually-defined patterns in~\cite{nature}. 
To ensure correctness of questions in  the test set, two doctors were asked to perform manual validation.

\section{Dataset Collection}
We develop a semi-automated pipeline to generate a pathology VQA dataset from pathology textbooks and online digital libraries. We manually check  the automatically-generated question-answer pairs to fix grammatical errors. The  automated pipeline consists of two steps: (1) extracting pathology images and  their captions from electronic pathology textbooks and the Pathology Education Informational Resource (PEIR) Digital Library\footnote{http://peir.path.uab.edu/library/index.php?/category/2} website; (2) generating questions-answer pairs from captions.

\subsection{Extracting Pathology Images  and Captions}

Given a pathology textbook that is in the PDF format and available online publicly, we use two third-party tools PyPDF2~\footnote{https://github.com/mstamy2/PyPDF2} and PDFMiner\footnote{https://github.com/pdfminer/pdfminer.six} to extract images and the associated captions therefrom. PyPDF2 provides APIs to access the ``Resources" object in each PDF page where  the ``XObject" gives information about images. 
PDFMiner allows one to obtain text along with its exact location in a page.
To extract image captions from text in each page, we use regular expressions to search for snippets with prefixes of ``Fig." or ``Figure" followed by figure numbers and caption texts. For a page containing multiple images, we order them based on their locations; the same for the captions. Images and locations are matched based on their order.  
 Given an online pathology digital library such as PEIR, we use two third-party tools Requests\footnote{https://requests.readthedocs.io/en/master/} and Beautiful Soup\footnote{https://www.crummy.com/software/BeautifulSoup/} to crawl images and the associated captions. Requests is an HTTP library built using Python and provides APIs to send HTTP/1.1 requests. Beautiful Soup generates the `http.parser' and can access the urls and tags of the images on the website pages. Given a set of urls, we use Requests to read website pages and use Beautiful Soup to find images under the targeted HTML tags including the Content Division element $\left< div\right>$, the unordered list element $\left<ul\right>$, and the $\left<li\right>$ element. Then we can download images with Requests and write their captions directly to local files. Given the extracted image-caption pairs, we perform post-processing including (1) removing images that are not pathology images, such as flow charts and portraits; (2) correcting erroneous matching between images and captions. \\

\subsection{Question Generation}

\begin{figure}[t]
 \centering 
 { 
    \includegraphics[width=\columnwidth]{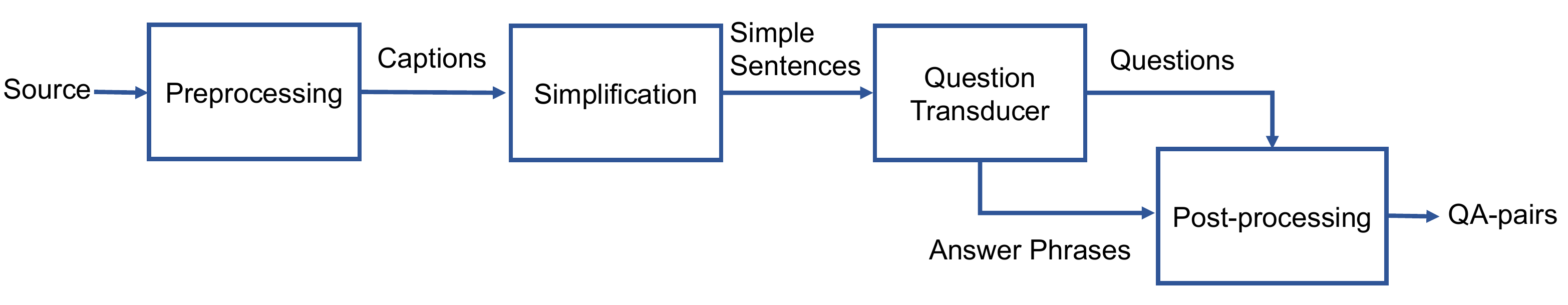}
    }
  \caption{The framework of generating questions from captions}
  \label{Q}
\end{figure}

\begin{figure}[htbp]
 \centering 
 { 
    \includegraphics[width =0.68\columnwidth]{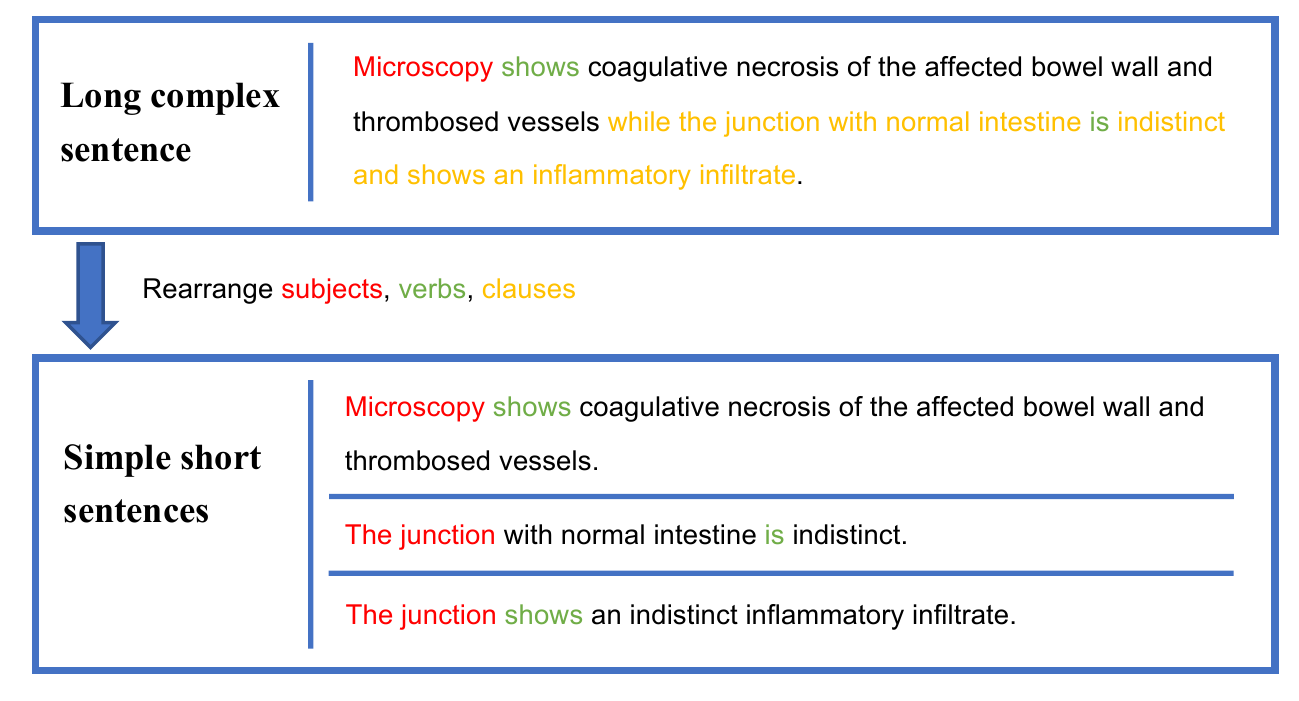}
    }
    \caption{Sentence simplification} 
  \label{simplification example}
\end{figure}

In this section, we discuss how to semi-automatically generate questions from captions. Figure \ref{Q} shows the overall framework.  We perform natural language processing of the captions using the Stanford CoreNLP~\cite{stanfordParser} toolkit, including sentence split, tokenization, part-of-speech (POS) tagging, named entity recognition (NER), constituent parsing, and dependency parsing. Many sentences are long, with complicated syntactic structures. We perform sentence simplification to break a long sentence into several short ones. Given the subjects, verbs, clauses, etc. labeled by POS tagging and syntactic parsing, we rearrange them using  the  rules proposed in~\cite{simplify_text,simplify_text2} to achieve simplification. 
Figure~\ref{simplification example} shows an example. 

Given the POS tags and named entities of the simplified sentences, we generate questions for them: including ``when"-type of questions  for  date and time entities and phrases such as ``in/during ... stage/period", ``before ...", and ``after ..."; ``how much/how many"-type of questions for words tagged as numbers; ``whose" questions for possessive pronouns (e.g., ``its", ``their"); ``where" questions for location entities and prepositional phrases starting with ``inner", ``within", ``on the right/left of"; ``how" questions for adjective words and phrases starting with ``using", ``via", ``with", and ``through", and  ``what" questions for the remaining noun phrases. Table~\ref{question_generate_example} shows an example for each type of questions.

\begin{table*}[t]
    \centering
\caption{Examples of generated questions for different types}
\scalebox{0.57}{
\begin{tabular}{|c|c|c|}
\hline
 Type& Original sentence& Question  \\
\hline 
\multirow{2}*{What} & \textbf{The end of the
long bone} is expanded  & What is expanded   \\
		~ & in the region of epiphysis. & in the region of epiphysis? \\
\hline
\multirow{2}*{Where}& The left ventricle is \textbf{on the lower right} & Where is the left ventricle \\
\multirow{-2}*{Where} & in this
apical four-chamber view of the heart. & in this apical four-chamber view of the heart?\\
\hline
When & \textbf{After 1 year of abstinence}, most scars are gone.& When are most scars gone? \\
\hline
How much/How many & \textbf{Two} multi-faceted gallstones are  present in the lumen. & How many multi-faceted gallstones are present in the lumen? \\
\hline
\multirow{2}*{Whose} & The tumor cells and \textbf{their} nuclei
are fairly uniform,& The tumor cells and whose nuclei are fairly uniform, \\
~ &  giving a monotonous appearance. &  giving a monotonous appearance?\\
\hline
 \multirow{2}*{How} & The trabecular bone
forming the marrow space shows trabeculae & How does the trabecular bone  \\
\multirow{-2}*{How} & \textbf{with osteoclastic activity at the margins}. & forming the marrow space show trabeculae?\\
\hline
\end{tabular}}
\label{question_generate_example}
\end{table*}

\begin{figure} [tbp]
  \centering 
 { 
    \includegraphics[width=0.75\columnwidth]{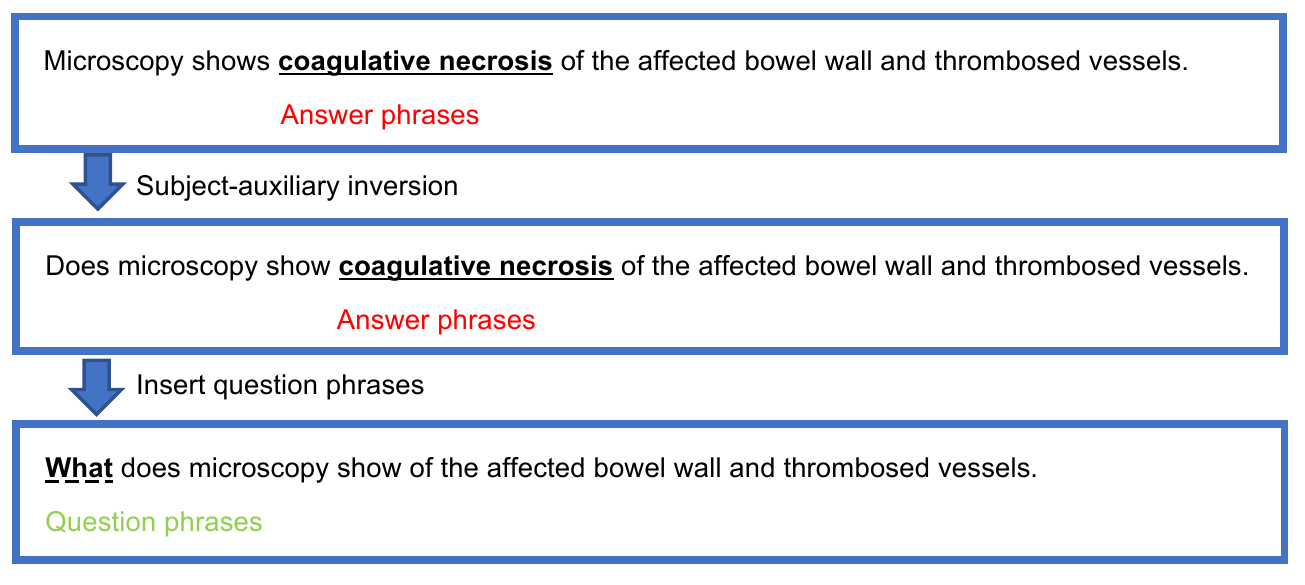}
    }
  \caption{ Implementation of syntactic transformation rules}
  \label{rules_example}
  \end{figure}

We use Tregex from Stanford CoreNLP tools \cite{stanford}, 
a tree query language including various relational operators based on the primitive relations of
immediate dominance and immediate precedence, to implement the rules~\cite{QG} for transforming declarative sentences (captions) into questions.
To reduce grammatical errors, we avoid generating questions on sentences with adverbial clauses such as ``chronic inflammation in the lung, showing all three characteristic  histologic  features". The question transducer mainly contains three steps. First, we perform the main verb decomposition based on the tense of the verb. For instance, we decompose ``shows" to ``does show". It is worth noting that for passive sentences with a structure of ``be+shown/presented/demonstrated", we keep their original forms rather than performing the verb decomposition. Second, we perform subject-auxiliary inversion. We invert the subject and the auxiliary verb in the declarative sentences to form the interrogative sentence. After the inversion, the binary ``yes/no" questions are generated, for instance, as shown in Figure~\ref{rules_example}, the sentence ``microscopy shows coagulative necrosis of the affected bowel wall and thrombosed vessels" is inverted to ``does microscopy show coagulative necrosis of the affected bowel wall and thrombosed vessels?". To generate questions whose answers are ``no", we randomly select a phrase with the  same POS  tagging from other captions to replace the  head words in the original question. For example, we replace ``coagulative necrosis" in the sentence ``does microscopy show coagulative necrosis of the affected bowel wall and thrombosed vessels" with other noun phrases. Third, we remove the target answer phrases and insert the question phrase obtained previously to generate open-ended  questions belonging to types of ``what", ``where", ``when", ``whose", ``how", and
``how much/how many" as shown in Table~\ref{question_generate_example}.
For instance, we transduce ``microscopy shows coagulative necrosis of the affected bowel wall and thrombosed vessels" to ``what of the affected bowel wall and thrombosed vessels does microscopy show?" as shown in Figure~\ref{rules_example}.

Given the automatically generated questions which may contain syntactic and semantic errors, we perform post-processing to fix those issues. We  manually proofread all questions to correct misspellings, syntactic errors, and semantic inconsistencies. The questions and answers are further cleaned by removing extra spaces and irrelevant symbols. Questions that are too  short or vague are removed. Articles appearing at the beginning of answers are stripped.
\section{Dataset Statistics}

\begin{table}[t]
    \centering
\caption{Statistics of our dataset}
\begin{tabular}{|l|c|c|c|}
\hline
& Maximum& Average& Minimum  \\
\hline 
\# questions per image& 14&6.6  & 1   \\
\hline
\# words per question& 28& 9.5& 3 \\
\hline
\# words per answer & 10& 2.5   &1  \\
\hline
\end{tabular}
\label{Statistics result}
\end{table}


Our PathVQA dataset consists of 32,799 question-answer pairs generated from 1,670 pathology images collected from two pathology textbooks: ``Textbook of Pathology" and ``Basic Pathology", and 3,328 pathology images collected from the PEIR\footnote{\url{http://peir.path.uab.edu/library/index.php?/category/2}} digital library.  Figure~\ref{illustrate} shows some examples. On average, each image has 6.6 questions. The maximum and minimum number of questions for a single image is 14 and 1 respectively. The average number of words per question and per answer is 9.5 and  2.5 respectively. Table~\ref{Statistics result} summarizes these statistics. There are 7 categories of questions: what, where, when, whose, how, how much/how many, and yes/no. Table~\ref{qresult} shows the number of questions and percentage of each category. The questions in the first 6 categories are open-ended: 16,465 in total and accounting for 50.2\% of all questions. The rest are close-ended ``yes/no" questions. The number of ``yes" and  ``no"  answers  are balanced, which is 8,145 and 8,189 respectively. The questions cover various aspects of the visual contents, including color, location, appearance, shape, etc. Such clinical diversity poses great  challenges for AI models to solve this pathology VQA problem.

\begin{minipage}{\textwidth}
\vspace{0.1in}
    \begin{minipage}[htbp]{0.7\textwidth}
        \centering
        \includegraphics[height=0.7\textwidth]{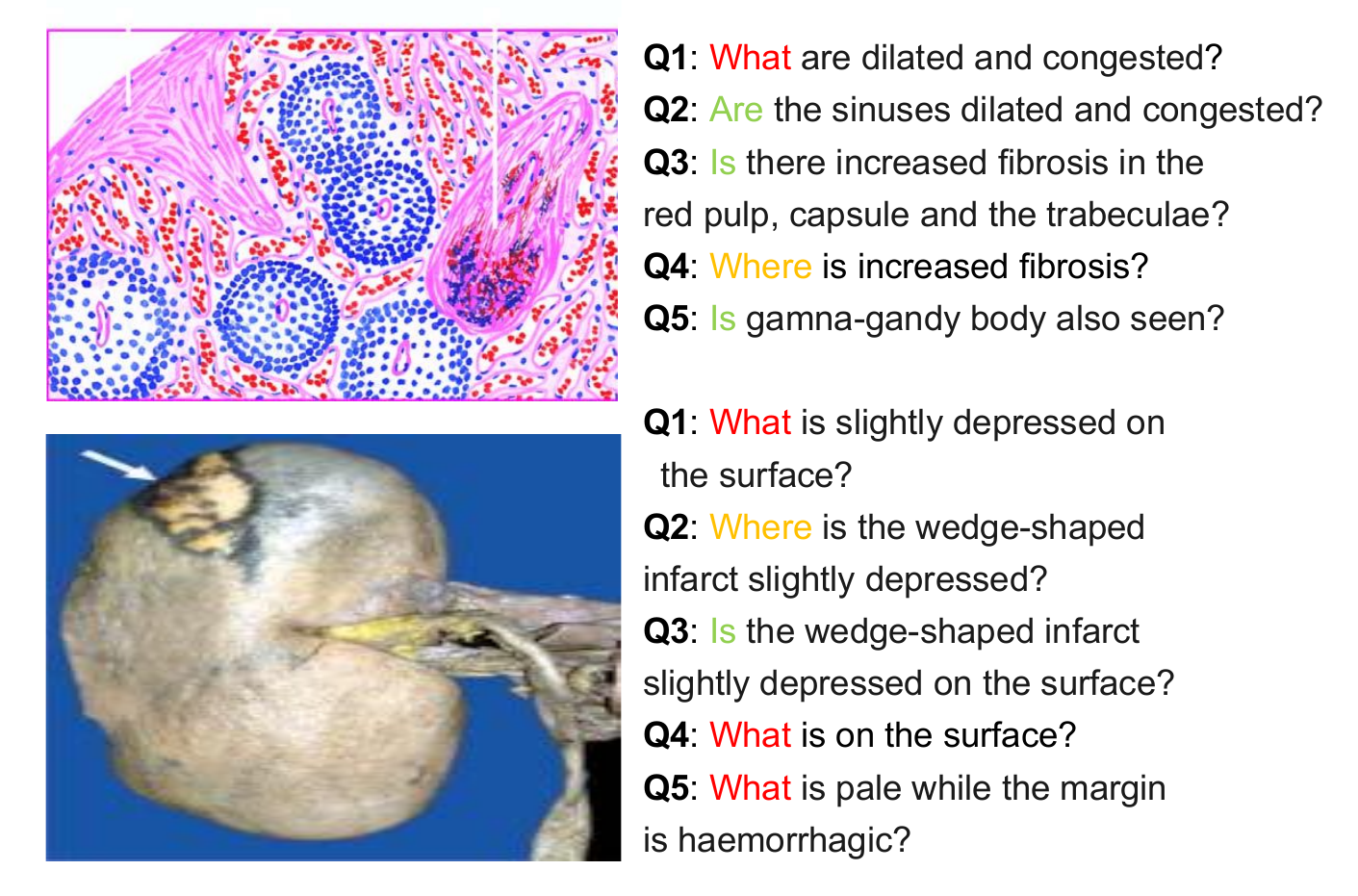}
        \makeatletter\def\@captype{figure}\makeatother\caption{Two exemplar images with semi-automatically generated questions. Both images have three types of questions: ``what", ``where", and ``yes/no".}
        \label{illustrate}
    \end{minipage}
    \begin{minipage}[h]{0.3\textwidth}
        \footnotesize
        \makeatletter\def\@captype{table}\makeatother\caption{Frequency of questions in different categories}
    \scalebox{0.85}{
    \begin{tabular}{|c|c|}
    \hline  
    \multirow{2}*{Question type} & Total number   \\
	\multirow{-2}*{Question type} & and percentage \\
    \hline
    Yes/No & 16,334 (49.8\%)\\
    \hline
    What&13,402 (40.9\%) \\
    \hline
    Where&1,268 (4.0\%) \\
    \hline                                                        
    How& 1,014 (3.0\%)\\
    \hline
    How much/How many& 294 (0.9\%)  \\
    \hline
    When&285 (0.9\%)\\
    \hline
    Whose& 202 (0.6\%)  \\
    \hline
    \end{tabular}}
        \label{qresult}   
    \end{minipage}
    \vspace{0.1in}
\end{minipage}

\begin{figure}[htbp]
 \centering 
 { 
    \includegraphics[width =1.0\columnwidth]{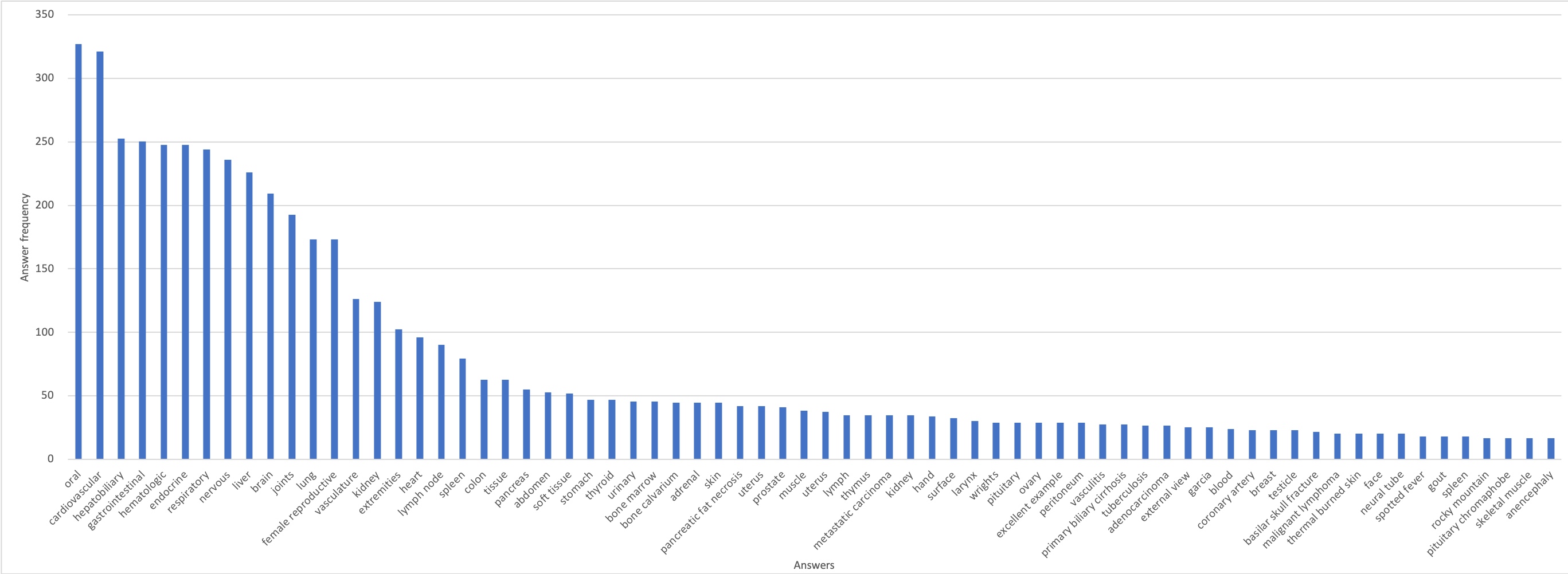}
    }
   \captionsetup{justification=centering}
  \caption{Frequencies of answers for open-ended questions} 
  \label{Frequency}
\end{figure}

Figure~\ref{Frequency} shows the  frequencies of different
answers for open-ended questions. The x-axis shows 70 most common answers and y-axis shows the frequency of each answer. As can be seen, the answer frequency has a long-tail distribution: a few answers have very high frequency while most answers have low frequency. Majority of answers have one or two words. 

To standardize the performance comparison on this dataset, we create an ``official" split. We randomly partition  the images along with the associated questions into a training set, validation set, and testing set with a ratio  of 0.5, 0.3, and 0.2. The statistics are summarized in Table~
\ref{data_in_each_set}. 
  \begin{table}[t]
    \centering
\caption{Statistics of data split}
\begin{tabular}{|c|c|c|c|}
\hline
& Training set& Validation set& Test set \\
\hline 
\# images& 2,499&1,499  & 1,000  \\
\hline
\# QA pairs&17,325 &9,462 &6,012 \\
\hline
\end{tabular}
\label{data_in_each_set}
\end{table}

\section{Benchmark VQA Performance}
In this section, we apply existing  well-established and state-of-the-art VQA methods to  our PathVQA dataset to obtain some baseline performance numbers for the research community to benchmark with. 

\subsection{Models}

We use three well-known VQA methods to generate the benchmark results.
\begin{itemize}
    \item \textbf{Method 1}: The method proposed in~\cite{BAN} uses a Gated Recurrent Unit (GRU)~\cite{GRU} recurrent network and a Faster R-CNN~\cite{RCNN} network to embed the question and the image.
 It learns bilinear attention distributions using the bilinear attention networks (BAN) and uses low rank approximation techniques to approximate the bilinear interaction between question embeddings and  image embeddings.

    \item \textbf{Method 2}: In~\cite{mcb}, a CNN is used to encode the image, and an LSTM~\cite{LSTM} network is used to encode the questions and answers. A multimodal compact bilinear pooling mechanism is proposed to match the image encoding and question encoding and an attention mechanism is leveraged to infer the answer.
    \item \textbf{Method 3}: The stacked attention network~\cite{SAN} embeds images and questions/answers using CNN and LSTM respectively and leverages a stacked attention mechanism to locate image regions that are relevant to answering the question. It queries the  image multiple times to progressively narrow the regions to be attended. 
\end{itemize}




\subsection{Experimental Settings}
Given the questions and answers, we perform standard pre-processing, including removing punctuation and stop words, tokenization, and converting to lower-cases. For question encoding and answer decoding, we create a vocabulary of 2200 words that have the highest frequencies. 
Data augmentation is applied to the images, including shifting, scaling, and shearing. In Method 1-3, we follow the original model configurations used in~\cite{BAN,mcb,SAN}, where the extraction of visual features are conducted using Faster R-CNN,  ResNet-152~\cite{resnet}, and VGGNet~\cite{VGG} respectively, with the Faster R-CNN pre-trained on Visual Genome~\cite{visualg} and the later two both pre-trained on ImageNet~\cite{imagenet}. Words in questions and answers are represented using GloVe~\cite{glove} vectors pre-trained on general-domain corpora such as Wikipedia, Twitter, etc. 
In Method 1, the dropout~\cite{dropout} rate for the linear mapping was set to 0.2 while for the classifier it was set to 0.5. The initial learning
rate was set to 0.005 with the Adamax optimizer~\cite{kingma2014adam} used. The batch size was set to 512. 
In Method 2, dropout was applied to the LSTM layers with a probability of 0.4. We set the feature dimension to 2048 in multimodal compact bilinear pooling. The optimizer  was Adam~\cite{kingma2014adam} with an initial learning rate of 0.0001 and a mini-batch size of 32.  
In Method 3, the  number of attention layers and LSTM layers were both set to 2 and the hidden dimensionality of
the LSTMs was set to 512. The weight parameters were learned using Stochastic Gradient Descent (SGD) with a momentum of 0.9, a learning rate of 0.1, and a mini-batch size of 100. As a comparison to Method 1 and Method 2, we change the image encoder in Method 3 to Faster R-CNN and ResNet-152 respectively. We refer to these two baseline models as Method 3 + Faster R-CNN and Method 3 + ResNet respectively.

\begin{minipage}{\textwidth}
\vspace{0.1in}
    \begin{minipage}[htbp]{0.55\textwidth}
        \centering
        \includegraphics[height=0.7\textwidth]{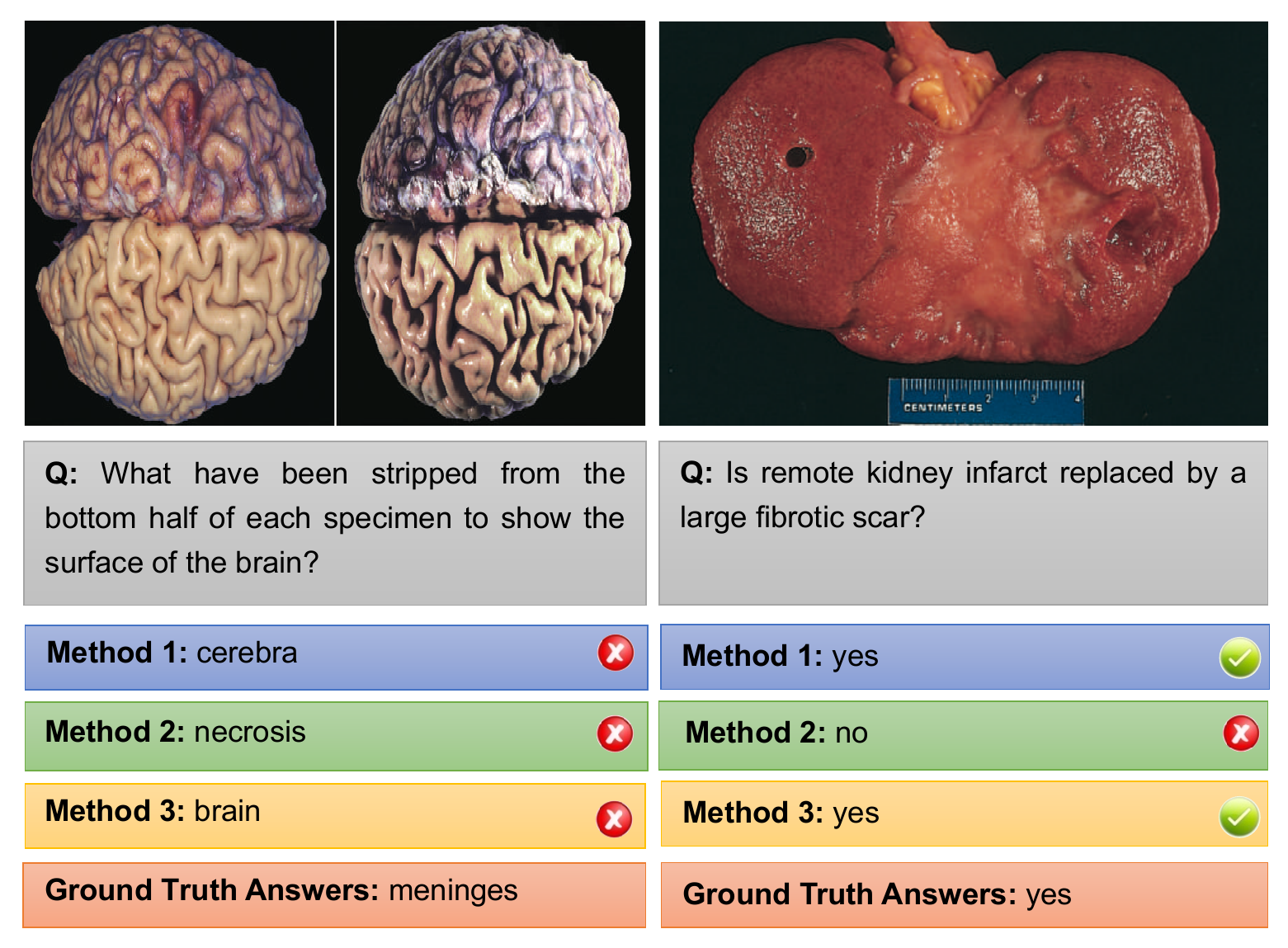}
        \makeatletter\def\@captype{figure}\makeatother\caption{Two qualitative examples of VQA}
        \label{illustration_example}
    \end{minipage}
    \begin{minipage}[h]{0.43\textwidth}
        \footnotesize
        \makeatletter\def\@captype{table}\makeatother\caption{Accuracy on ``yes/no" questions}
    \begin{tabular}{|c|c|}
    \hline   
     Method&Accuracy  (\%)\\
    \hline 
    Method 1 &68.2\\
    \hline
    Method 2&57.6\\
    \hline
    Method 3&59.4 \\
    \hline
    Method 3 + Faster R-CNN&62.0\\
    \hline
    Method 3 + ResNet&60.1\\ 
    \hline
    \end{tabular}
    \label{test result accuracy} 
    \end{minipage}
    \vspace{0.1in}
\end{minipage}

For ``yes/no" questions, we evaluate using accuracy. For open-ended questions, we evaluate using three metrics: (1) exact match~\cite{first}, which measures  the  percentage of inferred answers that match exactly with the ground-truth; (2) Macro-averaged F1~\cite{F1}, which measures the average overlap between the predicted answers and ground-truth, where the answers are treated as bag of tokens; (3) BLEU~\cite{bleu}, which measures the similarity of  predicted answers and ground-truth by matching $n$-grams.  

\subsection{Results}
Table~\ref{test result accuracy} shows the accuracy achieved by different methods on the ``yes/no" questions. All methods perform better than random guess (where the accuracy is 50\%). This indicates that this dataset is clinically meaningful, which allows VQA models to be learnable. 
Among Method 1-3, Method 1 performs the best. One primary reason is that it uses the bottom-up mechanism to propose candidate image regions and extract region-specific visual features. Typically the  answer is only relevant to a small region of the entire pathology image. Method 1 effectively localizes images regions that are most helpful in inferring the correct answer. This can be further verified by comparing Method 3 + Faster R-CNN with Method 3, where the former outperforms the latter. Method 3 + Faster R-CNN extracts region-specific features while Method 3 extracts holistic features of the entire image. Besides, the use of residual learning of attention and the superiority of bilinear attention over other co-attention approaches also contribute to the highest accuracy of Method 1.
Another observation is that Method 3 outperforms Method 2. This is because Method 3 utilizes multiple layers of attention to progressively learn where to attend, therefore achieving better performance than Method 2 which utilizes a single layer of attention. Method 3 + ResNet works better than Method 3, due to the reason that ResNet can extract better visual features than VGGNet. 

Table~\ref{test result} shows the exact match scores, F1, and BLEU-(1,2,3) scores on open-ended questions belonging to the following categories: what, where, how, whose, and when. As can be seen, these scores are low in general, which indicates that our dataset is very challenging for medical VQA. As a reference, we summarize the exact match scores achieved by these baseline methods on general-domain VQA datasets in  Table~\ref{result on existing dataset}. As can be seen, those numbers are much higher. The reasons that our dataset is so challenging lie in the following facts. First, most questions in our dataset are open-ended where the number of possible answers is $O(V^L)$, where $V$ is the vocabulary size and $L$ is the expected length of answers. This easily incurs the out-of-vocabulary issue, where the words in test examples may never occur in the training examples. 
Second, the size of our dataset is much smaller, compared with general domain VQA datasets.  More innovations of VQA models are needed to bridge the performance gap.
\begin{table}[t]
    \centering
\caption{BLEU-(1,2,3), exact match scores, and F1 on open-ended questions}
\begin{tabular}{|c|c|c|c|c|c|}
\hline   
 Method&\multicolumn{5}{c|}{Evaluation metric}    \\
\hline
& BLEU-1& BLEU-2& BLEU-3 &Exact match (\%) &F1 (\%)\\
\hline
Method 1 & 32.4& 22.8   &17.4& 2.9& 24.0\\
\hline 
Method 2& 13.3&9.5  & 6.8 &0.4 & 12.5\\
\hline
Method 3& 19.2& 17.9& 15.8 & 1.6&19.7\\
\hline
Method 3 + Faster R-CNN&24.7& 19.1& 16.5 & 1.9&21.2\\
\hline
Method 3 + ResNet&19.9& 18.0& 16.0 & 1.6&19.8\\
\hline
\end{tabular}
\label{test result}
\end{table}
\begin{table}[t]
\footnotesize
    \centering
\caption{Exact match (\%) scores on general-domain VQA datasets}
\begin{tabular}{|c|c|c|c|c|}
\hline   
Method&\multicolumn{4}{c|}{Dataset}    \\
\hline
&DAQUAR& VQA-real& VQA v2& COCO-QA  \\
\hline 
Method 1&\ding{55} & \ding{55} &  66.0&\ding{55} \\
\hline
Method 2&\ding{55} & 61.1& 66.5 &\ding{55}\\
\hline
Method 3 &45.5 &  58.7  &\ding{55}&61.6 \\
\hline
\end{tabular}
\label{result on existing dataset}
\end{table}
The exact match scores on open-ended questions are much lower than the  accuracy scores on ``yes/no" questions. This is not surprising since the number of candidate answers for open-ended questions is vast while that of ``yes/no" questions is only 2. Similar to the conclusions drawn from Table~\ref{test result}, bilinear attention based Method 1 achieves the best performance.
Method 3 works better  than  Method 2 by utilizing a stack of attention mechanisms.
 
  \begin{figure}[!htbp]
 \centering 
 { 
    \includegraphics[width =0.9\columnwidth]{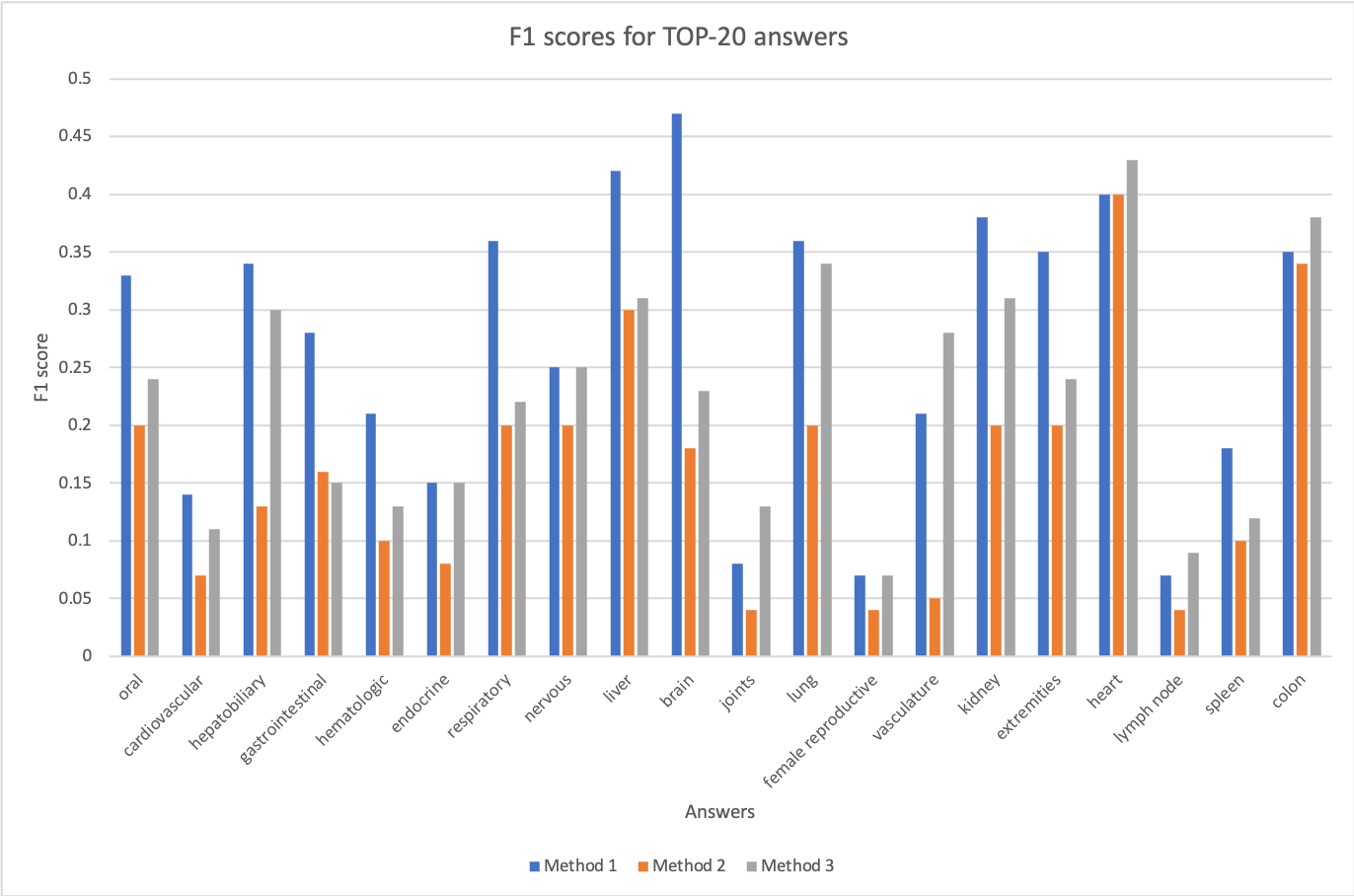}
    }
  \caption{F1 scores achieved on 20 most frequent answers by Method 1-3 } 
  \label{F1-score}
\end{figure}


Figure~\ref{F1-score} shows the individual F1 scores achieved on 20 most frequent answers by Method 1-3. As can be seen, Method 1 outperforms Method 2 and 3 on most answers. This is consistent with the results in Table \ref{test result accuracy} and \ref{test result}.

We show two qualitative examples of VQA results achieved by Method 1-3 
in Figure~\ref{illustration_example}. In the left example, all methods fail to give the correct answer since this answer is an infrequent one. But the three answers are semantically very relevant to the image, indicating that the models can learn something meaningful. In the right example, both Method 1 and Method 3 predict the answers correctly while Method 2 fails. This suggests that these two methods have certain advantages over Method 2 in that their effective attention mechanisms allow them to better recognize image regions of interest, which helps to give the correct answer.

\paragraph{Suggestions for model improvement} The visual feature extractors used in the baseline methods are pre-trained on general-domain images, which have a domain discrepancy with pathology images. One way to improve is to collect publicly available medical images (preferably pathology images) from textbooks, website, etc., whose domain is closer to that of the images in our dataset, then pre-train the CNNs using these medical images. Similarly for the word embeddings which were pre-trained on general-domain corpora, they may not be able to effectively capture the semantics relevant to pathology. To improve, we can pre-train the word embeddings on medical literature, such as medical textbooks, clinical guidelines, medical publications, etc.

\section{Conclusion and Future Works}
In this paper, towards the goal  of developing AI systems to pass the board-certificated examinations of the American Board  of Pathology and fostering research in medical visual question answering, we build a pathology VQA dataset that contains 32,799 question-answer pairs of 7 categories, generated from 4,998 images. Majority of questions in  our dataset are open-ended, posing great challenges for the medical VQA research. Our dataset is publicly available. 

For future studies, there are several aspects to improve. First, the questions in our dataset are not yet totally aligned with those in the ABP tests. In ABP test questions, each image is associated with a short text describing the medical history and demographics of the patient. These information are useful in determining the answers. To bridge this gap, we plan to create medical VQA datasets from the MedPix dataset where each image is associated with a caption and a text describing medical history and patient demographics. Second, in our current method, the creation of question/answer pairs from captions are mostly based on linguistic rules, which may not be diverse or robust enough. We plan to develop deep generative models that learn how to generate QA pairs from captions. Third, we plan to apply our automated pipeline to create VQA datasets for other types of medical images, such as radiology, ultrasound, CT scans, etc. Besides the board of pathology, other medical imaging domains have their own boards as well, organizing different types of board-certificated examinations. It would be interesting to build AI systems to pass those examinations as well. 

\clearpage
\bibliographystyle{unsrt}  
\bibliography{main}  


\end{document}